\documentclass{article}
\usepackage{PRIMEarxiv}

\usepackage{graphicx}
\usepackage[english]{babel}
\usepackage{times} 
\usepackage{url}
\usepackage{float}
\usepackage{verbatim}
\usepackage{mwe}
\usepackage{color}
\usepackage{hyperref}
\usepackage{amsmath}
\usepackage{bm}
\usepackage{multirow}
\usepackage[numbers]{natbib}

\usepackage{times}
\usepackage[utf8]{inputenc}

\usepackage{tabularx} 
\usepackage{ulem}
\usepackage[ruled,vlined,linesnumbered,noend]{algorithm2e}

\usepackage{caption}
\usepackage{subcaption}

\usepackage[misc]{ifsym}

\usepackage{xcolor}
\usepackage[colorinlistoftodos,prependcaption,textsize=tiny]{todonotes}

\emergencystretch=\maxdimen
\title{\textsc{SimpLex}: a lexical text simplification architecture}

\author{
    Ciprian-Octavian~Truică$^{1,2}$,
    Andrei-Ionuț~Stan$^2$,
    Elena-Simona~Apostol$^{1,2}$ \\
    $^1$ Department of Information Technology, Uppsala University, Uppsala, Sweden\\
    $^2$ Computer Science and Engineering Department, Faculty of Automatic Control and Computers, \\ University Politehnica of Bucharest, Bucharest, Romania \\
  \texttt{ciprian.truica@upb.ro, andrei\_ionut.stan@stud.acs.upb.ro, elena.apostol@upb.ro}
}

\begin{document}

\maketitle              

\begin{abstract}
  Text simplification (TS) is the process of generating easy-to-understand sentences from a given sentence or piece of text.
    The aim of TS is to reduce both the lexical (which refers to vocabulary complexity and meaning) and syntactic (which refers to the sentence structure) complexity of a given text or sentence without the loss of meaning or nuance.
    In this paper, we present \textsc{SimpLex}, a novel simplification architecture for generating simplified English sentences.
    To generate a simplified sentence, the proposed architecture uses either word embeddings (i.e., Word2Vec) and perplexity, or sentence transformers (i.e., BERT, RoBERTa, and GPT2) and cosine similarity.
    The solution is incorporated into a user-friendly and simple-to-use software.
    We evaluate our system using two metrics, i.e., SARI, and Perplexity Decrease.
    Experimentally, we observe that the transformer models outperform the other models in terms of the SARI score.
    However, in terms of Perplexity, the Word-Embeddings-based models achieve the biggest decrease.
    Thus, the main contributions of this paper are: 
    (1) We propose a new Word Embedding and Transformer based algorithm for text simplification;
    (2) We design \textsc{SimpLex} -- a modular novel text simplification system -- that can provide a baseline for further research; and
    (3) We perform an in-depth analysis of our solution and compare our results with two state-of-the-art models, i.e.,  LightLS~\cite{Glavas2015} and NTS-w2v~\cite{nisioi2017exploring}.
    We also make the code publicly available online.
\end{abstract}

\section{Introduction}
    
    Text simplification is a complex process that requires both a good knowledge of the language and a deep understanding of what constitutes a simple text.
    In spite of this, recent advances in Deep Learning and especially Natural Language Processing have enabled computer scientists to create systems that generate simplified text considered, even by human standards, to be of high quality~\cite{Grubisic2022}.
    The direction that has been explored in this field in the last years has been the usage of machine translation and Recurrent Neural Networks (RNNs) to automatically generate simplified sentences, to the detriment of manually crafted syntactic simplification rules and lexical substitutions~\cite{sikka2020survey}.
    Furthermore, with the new developments in the field of Machine and Deep Learning, new approaches for developing text simplification systems based on data-driven solution~\cite{AlvaManchego2020,Sauberli2020,Stodden2022} and few-shot learning~\cite{Jin2021,Qinyuan2021} have emerged.
    
    Text simplification has tremendous potential for helping specific groups of people in real-life situations~\cite{Gooding2022}.
    More specifically, text simplification can benefit people suffering from certain conditions, such as dyslexia, autism, and aphasia~\cite{Alarcon2021}.
    These individuals have trouble reading and understanding long sentences or sentences that contain complex words.
    Thus, automatic text simplification can open up more works to them.
    Text simplification also enables non-native speakers of a given language to understand pieces of text more easily without loss of meaning~\cite{AlThanyyan2022,sikka2020survey}.
    Current directions try to use (Neural) Machine Translation for this task~\cite{Erdem2022}, with a focus on encoder-decoder RNN based architectures~\cite{bahdanau2015neural,nisioi2017exploring}.
    These models can be further improved by using the attention mechanisms~\cite{luong2015effective} and different types of word embeddings, such as Word2Vec~\cite{Mikolov2013word2vec,nisioi2017exploring} or GloVe~\cite{Pennington2014,Stajner2017}.
    Although these approaches use machine translation methods for automatic language generation, they need a high level of explicit controllability, e.g., control the vocabulary size and words~\cite{nisioi2017exploring}, use special tokens to represent specific grammatical attributes~\cite{Martin2020}, employ domain specific dictionaries~\cite{Nassar2019}, use different word or sentence embeddings to find semantically and syntactical similar candidates for word~\cite{Stajner2017} and sentences~\cite{Kajiwara2016,Nishihara2019,Sjoblom2018}.

    As text simplification is crucial in helping specific groups of people with reading disabilities, and improving reading comprehension for non-native speakers as well as medical practitioners, teachers, and researchers, with this work, we try to answer the following research questions:
    \begin{itemize}
        \item[$Q_1$] Can we design a modular architecture for text simplification that uses both Word-based and Transformer-based embeddings?
        \item[$Q_2$] Can our architecture be used as a baseline for further research in the field?
    \end{itemize}
    
    To answer the two research questions, the main objectives of this paper are:
    \begin{itemize}
        \item[$O_1$] Propose a new easy to use modular text simplification architecture that can be used as a baseline solution for research in the field;
        \item[$O_2$] Show the efficiency of the proposed architecture by comparing it with state-of-the-art text simplification systems. 
    \end{itemize}
    
    In this paper, to address objective $O_1$, we propose \textsc{SimpLex} a simplification architecture that focuses on the lexical simplification of texts.
    \textsc{SimpLex} generates a simplified sentence that preserves the initial meaning of the original sentence.
    We propose two approaches for text simplification:
        \textit{(1)} Word Embeddings-based, and 
        \textit{(2)} Transformer-based.

    The Word Embedding-based approach uses the cosine similarity to select synonyms and then computes the perplexity score to determine the best simplified sentence.
    Word embeddings assign a static, non-changeable embedding to each word in the vocabulary.
    This approach can be problematic for a simplification system, as it deprives it of knowledge about the context in which the words are used.
    
    The Transformer-based approach selects synonyms based on transformer embeddings and then uses the cosine similarity to select the best simplified sentence.
    This approach generates embeddings both on the word and sentence level by taking a whole sentence and producing the embeddings.
    Thus, a single word can have different embeddings depending on the sentence and the context in which it is used.
    In this case, transformers prove to be more suited to choosing the best fitting word than word embeddings.
   
    We also provide a fully containerized server using docker and an easy-to-use UI that can serve as a tool for users to explore \textsc{SimpLex}.
    This is also an important contribution, as there are no commercial systems that provide automatic text simplification, mostly due to the fact that this is a very challenging task~\cite{Vstajner2021}.
    Furthermore, we make \textsc{SimpLex}'s source code publicly available online on GitHub at \url{https://github.com/elena-apostol/SimpLex}.

    To address objective $O_2$, we compare the results obtained by \textsc{SimpLex} with two state-of-the-art text simplification systems, i.e., LightLS~\cite{Glavas2015} and NTS-w2v~\cite{nisioi2017exploring}, on a well-established dataset in the community: WikiNet~\cite{Hwang-etal-2015}.

    The main contributions of this work are:
    \begin{itemize}
        \item[\textit{(1)}] The task of text simplification is a real-world problem that needs to be addressed. We try to propose a solution to this problem through \textsc{SimpLex}.
        \item[\textit{(2)}] We propose a novel architecture for text simplification that uses state-of-the-art NLP and NLU models for word complexity prediction and synonym-selection and ranking.
        \item[\textit{(3)}] We use both perplexity and cosine similarity to select the best synonym of a complex word and replace it with a simpler version without altering the meaning of the text.
        \item[\textit{(4)}] \textsc{SimpLex} uses a modular architecture, thus we can plug in other methods for synonym ranking (i.e., other word embeddings and other transformer models besides BERT, RoBERTa, and GPT2), even in other languages.
        \item[\textit{(5)}] We provide a fully functional implementation of \textsc{SimpLex}. As \textsc{SimpLex} can be deployed as a docker, this will help potential users of the software to easily use our solution.
        \item[\textit{(6)}] \textsc{SimpLex} can be used as a baseline for further research in the field of text simplification.
        \item[\textit{(7)}] Although evaluating human language complexity is tricky, we use a set of well-defined automatic methods to grant valuable insights into our results obtained with \textsc{SimpLex}.
    \end{itemize}

    The paper is structured as follows.
    In Section \ref{sec:related_work}, we discuss the current state-of-the-art in the field.
    In Section \ref{sec:metodology}, we present \textsc{SimpLex} and give a detailed view of each individual component.
    In Section \ref{sec:results}, we analyze the results achieved by our novel architecture and discuss some examples of sentence simplifications.
    Finally, in Section \ref{sec:conclusions}, we present the conclusions and future directions for further research and improvement.

\section{Related Work}~\label{sec:related_work}

Nassar et al. (2019)~\cite{nassar2019neural} discuss the two current approaches to the text simplification problem, namely the neural approach and the non-neural one.
They argue that, even though the neural approach is quite popular nowadays, a well-crafted rule-based architecture can outperform a neural-based system.
{\v{S}}tajner and Glava{\v{s}} (2017)~\cite{Stajner2017} propose a system that performs transformations, not only at the lexical and syntactic levels but also on the discourse level.
The system combines event-based simplification with lexical simplification and leads to significantly more content reduction within a sentence and within a text, managing to delete even whole sentences.
Furthermore, Zhong et al. (2020)~\cite{Zhong2020} observe that discourse level factors contribute to the challenging task of predicting sentence deletion for simplification.

Bahdanau et al. (2015)~\cite{bahdanau2015neural} introduce an encoder-decoder recurrent neural network (RNN) architectures for text simplification.
This model is improved by using the attention mechanism proposed by Luong et al. (2015)~\cite{luong2015effective} and different types of word embeddings, i.e., Word2Vec.
Nisioi et al. (2017)~\cite{nisioi2017exploring} utilize techniques from machine translation for text simplification and use a sequence-to-sequence encoder-decoder RNN-based architecture.
Both the encoder and the decoder have two long short-term memory (LSTM)~\cite{Hochreiter1997} layers of 500 hidden states of size 500.
A global attention mechanism combined with input feeding for the decoder is also employed.
Thus, the text simplification problem is viewed as a monolingual machine translation problem from a complex lexicon to a more simplified one.
The authors are able to improve the original system with respect to the metrics proposed in the original paper.
Surya et al. (2019)~\cite{Surya2019} propose a framework composed of a shared encoder and a pair of attentional-decoders assisted by discrimination-based losses and denoising.
The experimental results obtained on unlabeled texts collected from English Wikipedia show that the model performs lexical and syntactic text simplification.
Other approaches use document embeddings for aligning sentences.
Paun (2021)~\cite{paun2021} proposes a new unsupervised method for aligning text based on Doc2Vec embeddings and a new alignment algorithm capable of aligning texts at different levels.
Cumbicus-Pineda et al. (2021)~\cite{CumbicusPineda2021} propose a new edit-based system that deals with syntactical simplifications by employing edit operations at word level and graph convolutional networks to minimize the dependency of the sentence's structure. Using this approach, the system manages to extract the exact representation of syntax.

For lexical simplification, Maddela and Xu (2018)~\cite{maddela2018word} introduce a dataset of 15\,000 words labeled with a ranking from 1 to 6 by 11 human volunteers.
They also propose an architecture with both neural and non-neural elements to determine which words in a sentence are candidates for replacement, which candidates to generate, and, in the end, whether the new candidate sentences are indeed simpler than the original sentence.
Moreover, Konkol (2016)~\cite{konkol2016uwb} proposes a set of features for identifying complex words.
Using information such as the unigram and bigram probability of a given word, the number of sentences in which a word appears, and the WordNet Synset size of a word, the author is able to train an SVM (Support Vector Machines)~\cite{Cortes1995} model to predict if a word is complex or not.
Zhao et al. (2020)~\cite{Zhao2020} propose an asymmetric denoising autoencoder-based method for sentences with separate complexity to generate appropriate complex-simple pair. 
Alarcon et al. (2021)~\cite{Alarcon2021} propose a lexical simplification system for Spanish that 
\textit{(1)} identifies complex words using an SVM model, 
\textit{(2)} offers replacement candidates by employing a substitute generation approach using databases and a selection technique based on a pre-trained Word2Vec embedding model for Spanish, and
\textit{(3)} uses a pre-trained multilingual BERT model for word sense disambiguation.
Qiang et al. (2021)~\cite{Qiang2021} propose LSBert, a lexical simplification system based on pre-trained BERT to \textit{(1)} determine complex polysemous words and  \textit{(2)} improve the candidate selection.

Garbacea et al. (2021)~\cite{Garbacea2021} propose a compact pipeline for the simplification task. Their work focuses on the first two sub-tasks of the pipeline.
In the first step, they decide if a given text needs or not to be simplified by applying different traditional or deep classification models.
The second step is the explanation, where they highlight the part of the text that needs to be simplified.

Lin and Wan (2021)~\cite{Lin2021} propose a new model for Semantic Dependency Information guided Sentence Simplification (SDISS).
SDISS uses a sentence encoder to obtain contextual representations and a graph encoder to extract semantic dependencies.
Using these two encoders, the system manages to produce simplified sentences that are aware of the semantic dependencies between the words within a sentence.

Dehghan et al. (2022)~\cite{Dehghan2022} propose a new system (GRS) that combines generating and revision for unsupervised text simplification.
GRS uses BART-based~\cite{Lewis2020} paraphrasing and deletion as edit operations and a scoring function based on three metrics, i.e., Meaning Preservation, Linguistic Acceptability, and Simplicity, to search for the best simplification.
Devaraj et al. (2021)~\cite{Devaraj2021} also use paraphrasing to simplify medical texts.
The model is a transformer-based encoder-decoder that penalizes the decoder when producing 'jargon' terms using an unlikelihood loss function for augmentation.
The system is further studied by Devaraj et al. (2022)~\cite{Devaraj2022} who observe that the system yields errors at insertion, deletion, and substitution. 
Zhang et al. (2022)~\cite{Zhang2022} use a neural model for sentence deletion prediction to simplify text while maintaining the functional discourse structure.
The neural model uses a Bi-LSTM encoder-decoder structure with an attention layer between the encoder and decoder.
As input, the model receives BERT sentence representations.

\section{Methodology}~\label{sec:metodology}

\begin{figure*}[!ht]
        \centering
        \includegraphics[width=1\textwidth]{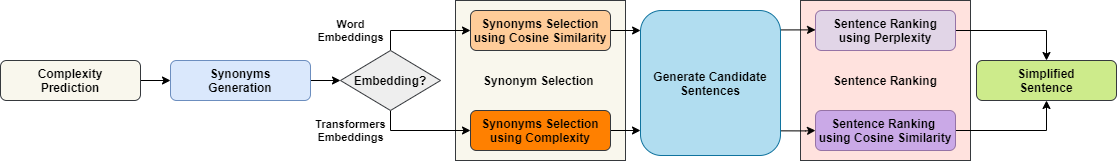}
        \caption{\textsc{SimpLex} architecture}
        \label{fig:my_label}
    \end{figure*}

\textsc{SimpLex}'s architecture is presented in Figure~\ref{fig:my_label} and it show the general overview of the different modules that are used by our solution to achieve the task of text simplification.
We describe in detail each of \textsc{SimpLex}'s module in the following subsections.

\textsc{SimpLex}'s input is a sentence, and the output is the simplified sentence. 

The first module, i.e., Complexity Predictions, determines if the words contained in a sentence are complex or not.
This module uses a Neural Network to predict the complexity of words within the input sentence.
Subsection~\ref{ssec:complex_pred} presents in detail this module.

The second module, i.e., Synonym Generation, is used to generate synonyms for the complex words determined by the Complexity Predictions module.
Subsection~\ref{ssec:canditate_generation} presents in detail this Synonym Generation module.

After determining which type of embedding to use, i.e., selecting between a word embedding approach (Word Embedding-based) or a transformer one (Transformer-based), the third module, i.e., Synonym Selection, is used to further refine the synonym list by choosing the synonyms that have the biggest similarity to the original word to be replaced.
Subsection~\ref{ssec:syn_selection} presents the Synonym Selection module.
The Word Embedding-based is presented in Subsubsection~\ref{sssec:syn_selection_cosine}, while the Transformer-based approach is presented in Subsubsection~\ref{sssec:syn_selection_cosine}.

The fourth module, i.e., Generate Candidate Sentences, is used to generate candidate sentences regardless of the embedding.
Subsection~\ref{ssec:canditate_generation} presents in detail the Generate Candidate Sentences module.

The fifth module, i.e., Sentence Ranking, is used to rank sentences and select the best sentence that is simpler than the original one and also keeps the meaning intact.
Subsection~\ref{ssec:sent_rnk} presents the Sentence Ranking module.
The perplexity ranking approach is presented in Subsubsection~\ref{sssec:sent_rnk_ppl}, while the cosine similarity ranking approach is presented in Subsubsection~\ref{sssec:sent_rnk_cos}.
The output of this module is the simplified sentence.


%

In Subsection~\ref{subsec:algorithm}, we present \textsc{SimpLex}'s algorithm which puts all these modules together in order to obtain the best simplified version for a given sentence. 

Finally, in Subsection~\ref{subsec:implementation}, we present the implementation details and user interface, and we provide the open source link to \textsc{SimpLex}'s source code.

\subsection{Complexity Prediction}\label{ssec:complex_pred}
    
Given a sentence, the Complexity Prediction module decides which words are the best candidates for a potential replacement.
Failure to select the right words or selecting too many words can lead to candidate sentences that remain the same or are more complex and obfuscated than the original ones.
To determine whether a word is complex or not, we build a Multi Layer Perceptron neural network that classifies the words into complex or simple classes.
We use the dataset from~\cite{maddela2018word} containing words ranked on a scale from 1 to 6.
We take the complexity rankings of the words and set the threshold of complexity at 3 to create two classes with words ranked from 1 to 3 labeled as simple and words ranked from 3 to 6 labeled as complex.
After re-labelling the words, we select the features for each word, basing our assessment on the work presented in~\cite{konkol2016uwb}.
Thus, we select the following 5 features:
\begin{itemize}
    \item[\textit{(1)}] \textit{Unigram probability of apparition in the selected language model}: in a language model, words that appear more often are, in general, more common, and thus, less complex.
    \item[\textit{(2)}] \textit{Number of sentences in which the word appears in the selected language model}: this is an important feature, as a word that is not necessarily simple can appear many times in a single sentence because of the fact that is, for example, the subject of the sentence.
    If a word appears in just a few sentences, it may be that the word is not as common as other words that appear in many sentences.
    \item[\textit{(3)}] \textit{Number of apparition of the word in the selected language model}: the same reasoning as the unigram probability of apparition.
    \item[\textit{(4)}] \textit{Word length}: more of a heuristic, but small words tend to be simpler than long words.
    \item[\textit{(5)}] \textit{WordNet Synset size of the word}: in WordNet~\cite{Miller1995}, a Synset of a word consists of words that express the same concept.
    The larger the Synset, the more meanings a word has.
    Thus, it can potentially be more complex.
\end{itemize}

To train our model, we need to choose a dataset to train our complexity prediction model.
The choice of the dataset is especially important, as it must reflect in a realistic manner the common words used in day-to-day speech.
For example, a dataset that contains academic discourse might not be a very inspired choice, as the words that are used frequently in this type of dataset might not be necessary simple or common words.
We choose the News Crawl dataset~\cite{Bojar2016} with news articles from 2017 as mass media employs easy-to-understand words in their writings.

\begin{figure}[!htbp]
    \centering
    \includegraphics[width=1\columnwidth]{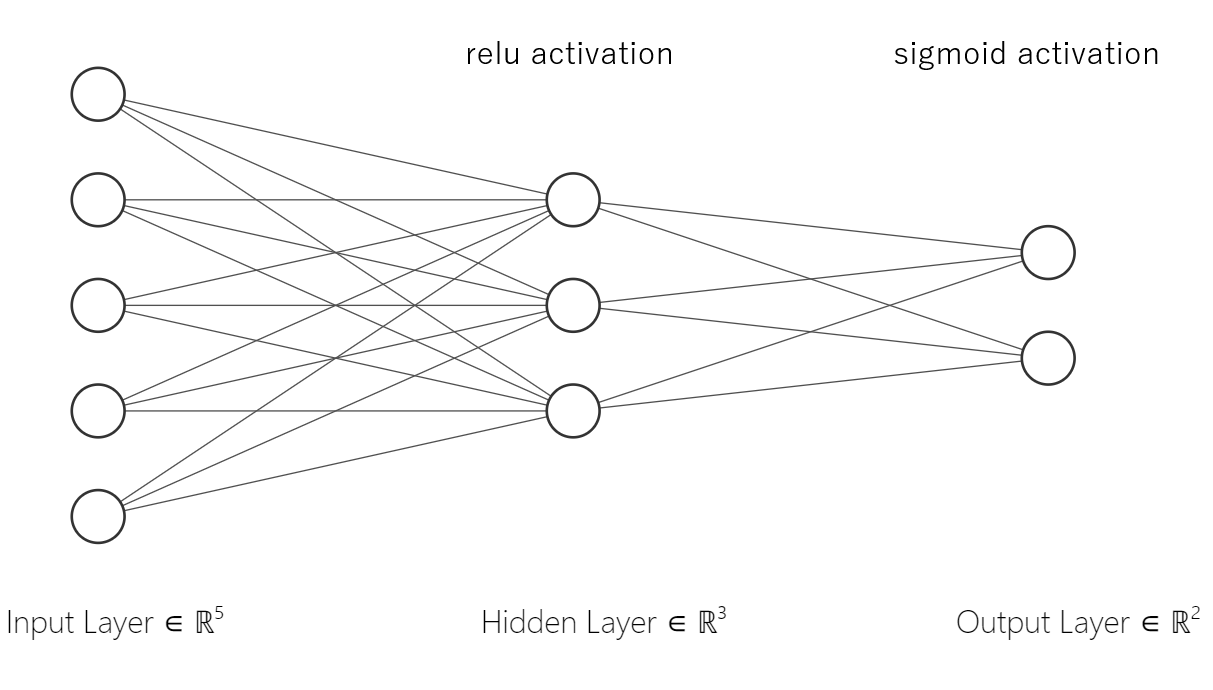}
    \caption{The neural network for complexity prediction}
    \label{fig:small-nn}
\end{figure}

We train a Multi Layer Perceptron (MLP) neural network with one hidden layer of size 3 (Figure~\ref{fig:small-nn}).
The neural network's input layer contains 5 units.
Each unit corresponds to one of the 5 features identified above.
The output is a one-hot-encoder vector that determines the probability of a word to be complex or simple, i.e, the vector $[1 \; 0]$ means the word is simple, while the vector $[0 \; 1]$ means the word is complex.
The hidden layer employs ReLU as the activation function.
The ReLU activation function used is defined as the positive part of its argument (Equation~\eqref{eq:relu}).
We choose this activation function as it solves the vanishing gradient problem~\cite{Glorot2011}.
The final layer contains 2 units for classifying the words into complex or simple.
These units employ the Sigmoid activation function (Equation~\eqref{eq:sigmoid}).
The Sigmoid function is used to predict probabilities that exist in the range $[0, 1]$, mapping the output of the model to the on-hot-encoder class representation.
The proposed neural network uses ADAM~\cite{Kingma2015} as the solver for weight optimization, along with a constant learning rate of 0.001.
The ADAM optimizer is a good choice for sparse gradients as it is invariant to their diagonal rescale~\cite{Kingma2015}.

\begin{equation}\label{eq:relu}
    f(x) = max(0, x)
\end{equation}
    
\begin{equation}\label{eq:sigmoid}
    f(x)=\frac{1}{1 + e^{-1}}
\end{equation}

\subsection{Synonym Generation}\label{ssec:syn_generation}
    
The Synonym Generation module builds a list of potential replacements for each candidate complex word.
At this stage of the simplification pipeline, it is not necessary to select the best candidates from among the synonyms, as there will be a specialized module for this task.
In the literature, two main directions have been explored regarding synonym generation: the thesaurus-based approach and the automatic approach.
In the automatic approach, pairs of substitutions are generated from aligned corpora of texts~\cite{sikka2020survey}.
By far, the most popular dataset has proven to be the aligned Wikipedia-Simplified dataset.
Yatskar et al. (2010)~\cite{yatskar2010sake} explored an approach in which they extracted pairs of simplifications from Simple Wikipedia edits made by users.
The thesaurus approach is much simpler, as it simply requires querying a source and extracting the result.
This has the added advantage that many reputable thesauri have been checked manually, and miss-matches are not very frequent.
The main problem with a thesaurus-based approach is that the context of the word is not known.
Thus, a word that manifests a high degree of polysemy can pose a problem for such a system, resulting in irrelevant matches.
Furthermore, the part of speech that the word has in a sentence must be taken into consideration: take, for example, the word 'looks' which can be used either as a verb (to look) or as a noun (appearance).
    
In the \textsc{SimpLex} architecture, we employ a thesaurus-based search, querying \url{https://www.synonym.com/}.
Prior to querying the thesaurus, we transform the word to its base form, the so-called lemmatization step.
In this way, we make sure that we have the best chance of obtaining the ideal candidate among the synonym list.
We also perform the following steps:
\begin{itemize}
    \item[\textit{(1)}] select only those words that have the same part of speech as the original word, and
    \item[\textit{(2)}] transform the remaining candidates to the original form required by the sentence.
\end{itemize}

The transformation operations include: declination of the verbs to the correct tense and person, pluralization for the nouns, morphological agreement, and comparative and superlative modes for the adjectives and adverbs.
    
After these steps are completed, we obtain a list of synonyms that can be immediately plugged in the original sentence in place of the original word and still create grammatically correct sentences.
In the following modules, the synonyms can be further refined by taking into consideration more complex aspects of the sentence.
    
\subsection{Synonym Selection}~\label{ssec:syn_selection}
    
    In this module, we aim to further refine the synonym list by choosing the synonyms that have the biggest similarity to the original word to be replaced.
    We use two distinct approaches, one based on word embeddings cosine similarity (Word Embedding-based) and one based on transformer embeddings (Transformer-based).

\subsubsection{Synonym Selection using Cosine Similarity.}~\label{sssec:syn_selection_cosine}
For the Word Embedding-based approach, we compute the cosine similarities (Equation~\eqref{eq:cossim}) between the embedding $\mathbf{s} = \{s_1, \dots, s_n \}$ of synonym $syn$ and the embedding $\mathbf{w} = \{w_1, \dots, w_n\}$ of the target word $w$.
We keep only the synonyms that have a cosine similarity above the average of all the similarities.

\begin{equation}\label{eq:cossim}
    cos(\mathbf{w},\mathbf{{s}}) = \frac{\sum_{i=1}^{n}{w_{i}{s}_{i}}}{\sqrt{\sum_{i=1}^{n}{w_{i}^{2}}} \sqrt{\sum_{i=1}^{n}{{s}_{i}^{2}}}}
\end{equation}    

Thus, we make sure that the new candidate words are indeed suitable replacements for the target word.
We can now remark that using word embeddings, the initial thesaurus search can be bypassed if we consider the top-$k$ best words with respect to their cosine similarities to the original word.
As it turns out, the thesaurus search is indeed necessary.
Embedding models are able to determine correct relations between words, such as synonymy, but are prone to labeling antonyms as related words~\cite{mikolov2013linguistic} (for example, the words 'king' and 'queen' will be branded as similar, even though they are not real synonyms).
Thus, we can end up in a situation in which the words are not really synonyms but related words with similar meanings.
Using the thesaurus search, we ensure that the selected words are already good matches and just keep those that are the most similar to the candidate word.

\subsubsection{Synonym Selection using Complexity.}~\label{sssec:syn_selection_complexity}
For the Transformers-based approach, we do not need to check if the candidate synonym is similar to the target word because this is automatically done when a sentence embedding is created.
Thus, we only check if the candidate synonym is indeed simpler than the word we want to replace.
Because the target word is already marked as complex, it is sufficient for the synonym to be predicted as a simple word.
Furthermore, we keep only those synonyms that are being predicted by the complexity prediction Multi Layer Perceptron neural network as simple.

\subsection{Generate Candidate Sentences}~\label{ssec:canditate_generation}

    Regardless of the embedding, this module generates candidate sentences.
    Given a sentence $S = \{w_1, \dots, w_n\}$, a word $w$ at position $ k \in \overline{1, n}$, and a list of synonyms $syns$, we obtain a $candidate$ sentence by replacing $w$ with a synonym $syn \in syns$ without changing any of the existing words that are on the left ($w_l, 0 \leq l < k$) or write ($w_r,  k < r \leq n$) of word $w$.
    The candidate sentences consist of the original sentence in which the complex word detected is replaced by a synonym.
    We also perform the transform step from the Synonym Selection module.

\subsection{Sentence Ranking}~\label{ssec:sent_rnk}
   
    After all the candidate sentences are generated, this module chooses the best sentence that is both simpler than the original one and also keeps the meaning intact.
    
    \subsubsection{Sentence Ranking using Perplexity.}~\label{sssec:sent_rnk_ppl}
    For the Word Embedding-based, we use perplexity as the metric for ranking sentences.
    We define the perplexity of a sentence $S = \{w_1,\dots, w_n\}$ with respect to a given language model as $PP(S)$ (Equation~\eqref{eq:perplexity}).
    The rewritten form of perplexity for unigrams (under the assumption that all words are independent) is $PP_{1}(S)$ (Equaition~\eqref{eq:perplexity_1gram}).
    While for a bigram, under the first order Markov assumption, is $PP_{2}(S)$ (Equation~\eqref{eq:perplexity_2gram}).
    For both formulas, $p(w) = \frac{f_w}{|V|}$ and $p(w | v) = \frac{f_{v,w}}{f_v}$, 
    where $f_w$, $f_v$, $f_{v,w}$ are the frequency of $w$, $v$, and the co-occurrence of $w$ and $v$ respectively, and $|V|$ is the size of the vocabulary.

    \begin{equation}\label{eq:perplexity}
        PP(S) = p(w_1, w_2, ..., w_n)^\frac{1}{n}
    \end{equation}
    
    \begin{equation}\label{eq:perplexity_1gram}
        PP_{1}(S) = (p(w_1) p(w_2) \dots p(w_n))^\frac{1}{n}
    \end{equation}
    
    \begin{equation}\label{eq:perplexity_2gram}
        PP_{2}(S) = (p(w_1) p(w_2|w_1) \dots p(w_n|w_{n-1}))^\frac{1}{n}
    \end{equation}

    Because of the small occurrence probabilities and their multiplication, the values can become very small, and numerical stability problems can occur.
    Thus, we can rewrite the relations for perplexity using Equation~\eqref{eq:perplexity_1gram_rw} and~\eqref{eq:perplexity_2gram_rw} for unigrams and bigrams, respectively.
    
    \begin{equation}\label{eq:perplexity_1gram_rw}
        PP_{1}(S) = 2^{-\frac{1}{n}\sum_{i=1}^{n} log(p(w_i))}
    \end{equation}
    
    \begin{equation}\label{eq:perplexity_2gram_rw}
        PP_{2}(S) = 2^{-\frac{1}{n}(log(p(w_1)) + \sum_{i=2}^{n} log(p(w_i|w_{i-1})))}
    \end{equation}
    
    In our sentence ranking, we use a linear combination of the two perplexity metrics.
    Equation~\eqref{eq:score} presents this score, which uses $\varphi$ to minimize the impact of the bigrams.
    Rewritten with numerical stability in mind, the final scoring function is presented in Equation~\eqref{eq:score_rw}.
    
    \begin{equation}\label{eq:score}
        PP_{Score}(S) = (1 - \varphi) \cdot PP_{1}(S) + \varphi \cdot PP_{2}(S)
    \end{equation}
    
    \begin{equation}\label{eq:score_rw}
        \begin{split}
            PP_{Score}(S) & = (1 - \varphi) \cdot 2^{-\frac{1}{n}\sum_{i=1}^{n} log(p(w_i))} + \\
                            & \varphi \cdot 2^{-\frac{1}{n}(log(p(w_1)) + \sum_{i=2}^{n} log(p(w_i|w_{i-1})))}    
        \end{split}
    \end{equation}

    It should be noted that this approach ensures that sentences that are more probable with respect to the chosen language model are selected.
    This is generally desirable in a text simplification tool, but one major drawback of this approach is that it does not have any clue about the context of the words.
    This can become problematic in the case of polysemy.
    Take, for example, the noun 'spider', which can mean the arachnid or, more rarely, an additional rest cue used in billiards.
    A simplification system that does not know context might interpret the word 'spider' in the sentence 'The English player used the spider to execute the shot' as referring to the arachnid.
    This is why it is desirable to introduce embeddings that are context-aware.
    
    \subsubsection{Sentence Ranking using Cosine Similarity.}~\label{sssec:sent_rnk_cos}
    For the Transformer-based approach, we exploit the \textit{sentence-level embeddings} created by the transformer embedding.
    The advantage of transformers over word embeddings is that the embeddings for a given word can change depending on the context in which that particular word is used in a sentence, while word embeddings are static embeddings.
    
    After we have sentence embeddings, we can select the candidate sentence that has the biggest cosine similarity to the original sentence.
    These cosine similarities have, in general, high values above 0.9.
    This is because only a word has been replaced, and the other words remain the same.
    Thus, the sentence remains mainly unchanged.
    This is not an issue, as differences are still visible between the words that fit the sentence and keep the context, and words that change the context or meaning.
    It is interesting to note that the approach using transformers will not necessarily produce a new sentence with the lowest perplexity, but this is not always a bad thing, as in some cases, the meaning cannot be fully preserved by just looking for the smallest perplexity.

\subsection{\textsc{SimpLex} Algorithm}\label{subsec:algorithm}

    Algorithm~\ref{alg:simplex} presents the \textsc{SimpLex} pseudocode.
    Given a sentence $S$ (Line~\ref{algo:l1}), we test if each word $w$ is complex or not (Line~\ref{algo:l2}) using the methodology presented in Subsection~\ref{ssec:complex_pred}.
    If the word $w$ is complex, then we generate the list of its synonyms $syns$ (Line~\ref{algo:l3})
    using the methodology presented in Subsection~\ref{ssec:syn_generation}.
    We check the synonym complexity and update the synonyms list removing all the words that are not deem complex by the proposed Multi Layer Perceptron complexity prediction classifier (Subsection~\ref{ssec:complex_pred}) (Lines~\ref{algo:l4} to~\ref{algo:l5}).

    \normalem
    \begin{algorithm}[!htbp]
    \small
    \SetKwInput{KwInput}{Input}
    \SetKwInput{KwOutput}{Output}
    \SetAlgoNoLine
    \SetAlgoNoEnd
    \DontPrintSemicolon
    \SetInd{0.2em}{0.7em}
    \newcommand\mycommfont[1]{\footnotesize\ttfamily\textcolor{blue}{#1}}
    \SetCommentSty{mycommfont}
    \KwInput{A complex sentence $S$}
    \KwOutput{A simplified sentence $S'$}
    \ForEach{word $w$ in $S$} {\label{algo:l1}
        \tcp{Determine $w$ complexity}
        \If {$w$ is a complex word} {\label{algo:l2}
            \tcp{Get all synonyms $syn$ for the word $w$}
            $syns \gets \{ syn \mid syn \sim w \}$\;\label{algo:l3}
            
            \BlankLine
            \tcp{Check the synonym complexity}
                \For{$syn \in syns$} {\label{algo:l4}
                    \If{$syn$ is not a simple word}{
                        $ syns \gets syns \setminus \{ syn \} $\;
                    }
                }\label{algo:l5}

            \If{use \textsc{Word Embeddings}}{\label{algo:we_0}
            
                \tcp{Compute the average $cos$}
                ${cos}_{avg} \gets 0$\;\label{algo:we_1}
                $\mathbf{w} \gets WordEmbedding(w)$\;
                \ForEach{$syn \in syns$}{
                    $\mathbf{s} \gets WordEmbedding(syn)$\;
                    ${cos}_{avg} \gets{cos}_{avg} + cos(\mathbf{w}, \mathbf{s})$\;
                }
                ${cos}_{avg} \gets \frac{{cos}_{avg}}{|syns|}$\;\label{algo:we_2}

                \BlankLine
                \tcp{Keep only synonyms with $cos \ge {cos}_{avg}$  }
                $syns' \gets \emptyset$\;\label{algo:we_3}
                \ForEach{$syn \in syns$}{
                    $\mathbf{s} \gets WordEmbedding(syn)$\;
                    \If{${cos}_{avg} \ge cos(\mathbf{w}, \mathbf{s})$}{
                        $syns \gets syns \setminus \{ syn \}$\;
                    }
                }\label{algo:we_4}

                \BlankLine
                \tcp{Generate candidate sentences}
                $candidates  \gets \emptyset$\;\label{algo:we_5}
                \ForEach{$syn \in syns$}{
                    \tcp{Replace word $w$ with $syn$ in $S$}
                    $candidate \gets$ \parbox[t]{.5\linewidth}{%
                        $\{w_l | w_l \in S \wedge 0 \leq l < k \} \cup \{ syn \} \cup$ \\ 
                        $\{w_r | w_r \in S \wedge k < r \leq n \}$
                    \;}
                    
                    $candidates \gets candidates \cup \{ candidate \}$\;
                }\label{algo:we_6}
                
                \BlankLine
                \tcp{Get the candidate with the lowest perplexity}
                $PP_{min} \gets \infty$\;\label{algo:we_7}
                \ForEach{$candidate \in candidates$}{

                    \If{$PP_{Score}(candidate) < PP_{min}$}{
                        $PP_{min} \gets PP_{Score}(candidate)$\;
                        $S' \gets candidate$\;
                    }
                }\label{algo:we_8}
            }
            \ElseIf{use \textsc{Transformers Embeddings}}{\label{algo:te_0}
                
                \tcp{Generate candidate sentences}
                $candidates  \gets \emptyset$\;\label{algo:te_3}
                \ForEach{$syn \in syns$}{
                    \tcp{Replace word $w$ with $syn$ in $S$}
                    $candidate \gets$ \parbox[t]{.5\linewidth}{%
                        $\{w_l | w_l \in S \wedge 0 \leq l < k \} \cup \{ syn \} \cup$ \\ 
                        $\{w_r | w_r \in S \wedge k < r \leq n \}$
                    \;}
                    
                    $candidates \gets candidates \cup \{ candidate \}$\;
                }\label{algo:te_4}
                
                \BlankLine
                \tcp{Get candidate with the highest $cos$}
                $cos_{max} \gets - \infty$\;\label{algo:te_5}
                $\mathbf{S} \gets TransformerEmbedding(S)$\;
                \ForEach{$candidate \in candidates$}{
                    $\mathbf{C} \gets TransformerEmbedding(candidate)$\;
                    \If{$cos(\mathbf{S}, \mathbf{C}) > cos_{max}$}{
                        $cos_{max} \gets cos(\mathbf{S}, \mathbf{C})$\;
                        $S' \gets candidate$\;
                    }
                }\label{algo:te_6}
                \tcp{Compute the perplexity for $S'$}
                $PP_{S'} \gets PP_{Score}(S')$\;\label{algo:te_7}
            }
        }
    }
    \Return{$S'$}\;\label{algo:l6}
    \caption{\textsc{SimpLex} algorithm}
    \label{alg:simplex}
    \end{algorithm}
    \ULforem 
    
    For the Word Embedding-based approach (Line~\ref{algo:we_0}), we perform the following steps as discussed in Subsection~\ref{sssec:syn_selection_cosine}:
    \begin{itemize}
        \item[\textit{(1)}] Compute the average cosine similarity between word embedding $\mathbf{w}$ of the $w$ and each embedding $\mathbf{s}$ of synonyms $syn \in syns$ (Lines~\ref{algo:we_1} to~\ref{algo:we_2});
        \item[\textit{(2)}] Keep only the synonyms that have a cosine similarity that is larger or equal to the average (Lines~\ref{algo:we_3} to~\ref{algo:we_4});
        \item[\textit{(3)}] Generate all candidate sentences $candidates$ by replacing $w$ with its synonyms $syn$ without changing any of the existing words position on the left ($w_l$) or right ($w_r$) of $w$ (Lines~\ref{algo:we_5} to~\ref{algo:we_6}) as discussed in Subsection~\ref{ssec:canditate_generation};
        \item[\textit{(4)}] Extract the candidate sentence $S'$ with the lowest perplexity score (Lines~\ref{algo:we_7} to~\ref{algo:we_8}) using Equation~\eqref{eq:score_rw} from Subsection~\ref{sssec:sent_rnk_ppl}.
    \end{itemize}
  
    When using Transformer-based approach (Line~\ref{algo:te_0}), the following steps are performed as discussed in Subsection~\ref{sssec:syn_selection_complexity}:
    \begin{itemize}
        \item[\textit{(1)}] Generate all candidate sentences $candidates$ using the same approach as when using word embeddings (Lines~\ref{algo:te_3} to~\ref{algo:te_4}) as discussed in Subsection~\ref{ssec:canditate_generation};.
        \item[\textit{(2)}] Use sentence embedding and extract the candidate sentence $S'$ with the highest cosine similarity to $S$ (Lines~\ref{algo:te_5} to~\ref{algo:te_6}) as presented in Subsection~\ref{sssec:sent_rnk_cos}.
        \item[\textit{(3)}] Compute the perplexity of $S'$ (Lines~\ref{algo:te_5} to~\ref{algo:te_7}) using Equation~\eqref{eq:score_rw} from Subsection~\ref{sssec:sent_rnk_ppl}.
    \end{itemize}

    Regardless of the used embedding, the algorithm returns the new simplified sentence $S'$ (Line~\ref{algo:l6}).
    We note that in our initial experiments, we used for the Word Embedding-based approach the cosine similarity as a metric for extracting the simplified sentence, but the results were worse than when using the perplexity score.
    Furthermore, in our Transformer-based approach, we used the perplexity score to extract the best simplified sentence, and we observed that this approach yields worse results than when using the cosine similarity, as can be seen in Section~\ref{sec:results}.

\subsection{Implementation and user interface}\label{subsec:implementation}

\textsc{SimpLex} is implemented in \textit{Python v3.7}.
We use \textit{NLTK}\footnote{\url{https://www.nltk.org/}}~\cite{bird2009natural} to preprocess the texts.
For the complexity prediction module, we employ the Multi-Layer Perceptron model from \textit{Scikit-Learn}\footnote{\url{https://scikit-learn.org/stable/}}~\cite{scikit-learn}.
To generate candidate synonyms for the complex words, we use the \textit{PyDictionary}\footnote{\url{https://github.com/geekpradd/PyDictionary}}~\cite{PyDictionary} library.
The morphological changes are made using \textit{pyinflect}\footnote{\url{https://github.com/bjascob/pyInflect}}~\cite{PyInflect} and \textit{pattern}~\cite{pattern} libraries.
For loading word embeddings, we use \textit{Gensim}\footnote{\url{https://radimrehurek.com/gensim/}}~\cite{rehurek_lrec}, while for the transformer embedding, we employ \textit{simpletransformers}\footnote{\url{https://simpletransformers.ai/}}~\cite{simpletransformers}.

\begin{figure*}[!htbp]
    \centering
    \includegraphics[width=1\columnwidth]{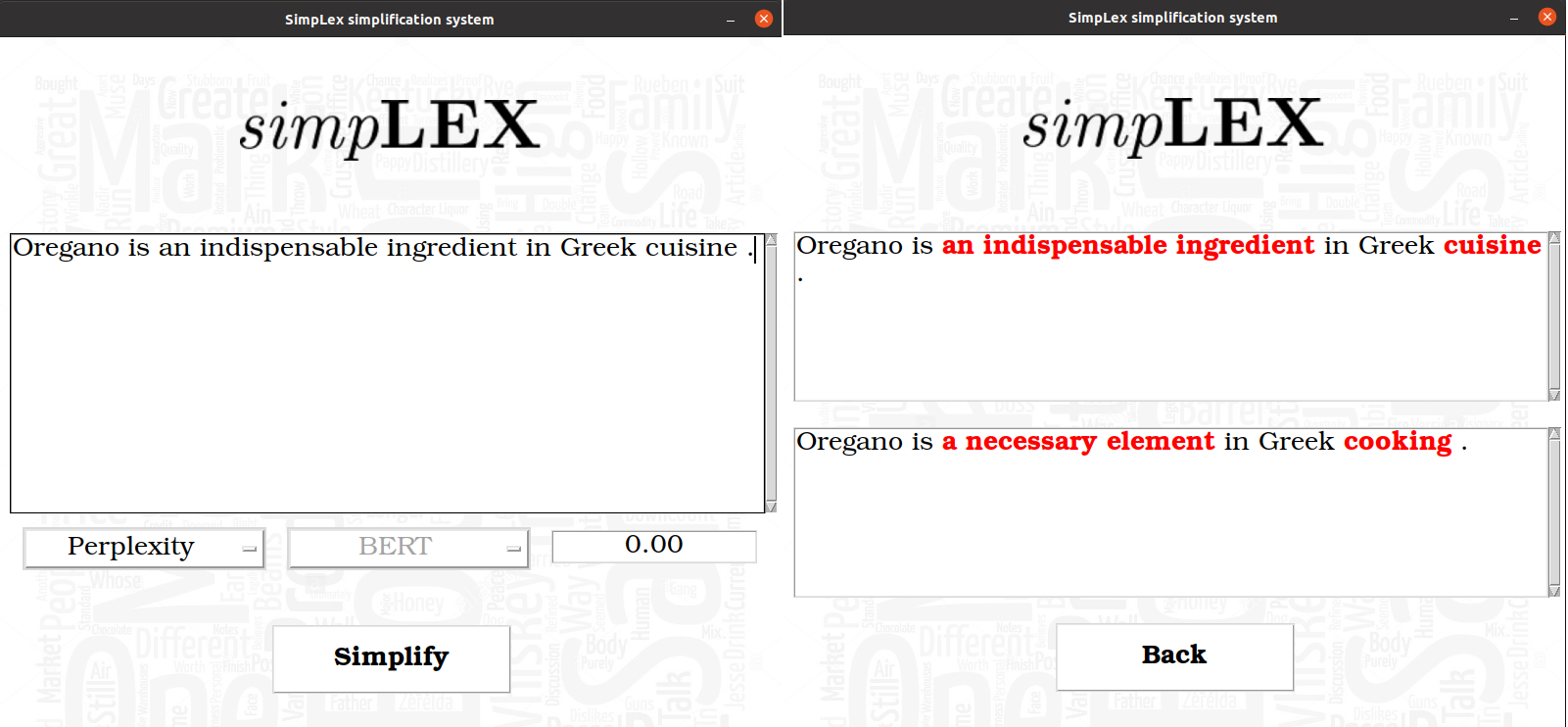}
    \caption{The interface for the \textsc{SimpLex} system}
    \label{fig:gui}
\end{figure*}

\textsc{SimpLex} also provides a simple-to-use and friendly user interface and full docker containerization~\cite{Merkel2014}.
In this way, a docker image is created from the Python official image and then the necessary third-party libraries are installed, along with the source code.
After this, a basic REST API server is launched in the new container.
\textsc{SimpLex} presents the user with a screen in which the sentence is introduced, and the simplification parameters are chosen (Figure~\ref{fig:gui}).
The interface provides a quick and easy way of visualizing lexical changes made by our system.

The code is publicly available on GitHub at \url{https://github.com/elena-apostol/SimpLex}.

\section{Results}~\label{sec:results}

In this section, we present the experimental evaluation of \textsc{SimpLex} and discuss the results.

\subsection{Datasets}
    
The dataset used for the testing of the simplification tool is WikiNet~\cite{Hwang-etal-2015} and consists of a set of 100 English sentences taken from the Wikipedia corpus.
The dataset contains alignments considered 'good' and 'partial good'.
To evaluate the Complexity prediction module, we employ other two datasets:
\begin{itemize}
    \item[\textit{(1)}] the Complexity Ranking dataset~\cite{maddela2018word}, and
    \item[\textit{(2)}] the news corpus from News Crawl dataset~\cite{Bojar2016} with news articles from 2017.
\end{itemize}

The Complexity Ranking dataset is a human-rated word complexity lexicon of 15\,000 English words.
The News Crawl dataset contains about 3.7 million sentences from all types of news, giving us a balanced and realistic look at the common words used in day-to-day speech.
    
\subsection{Evaluation metrics}
   
For the text simplification task, the evaluation is not as straightforward as for other machine learning tasks.
This is because what is considered simple text is a very subjective topic, as it depends on the reader's experience with the language, education, etc.
Nevertheless, researchers have been able to come up with a few interesting and efficient solutions to quantify the simplicity of a piece of text.

The two metrics that we choose for the evaluation of the simplification system are: SARI~\cite{xu2016optimizing} and Perplexity Decrease~\cite{Zhao2020}.
SARI quantifies not only replacement correctness but also syntactical simplifications.
It rewards both word replacement and eliminations, resulting in a more flexible evaluation for a simplification system.
The Perplexity Decrease metric tracks how the perplexity changes when lexical simplifications are applied to the original text.
It is desirable to generate sentences with a smaller perplexity so that they are more likely to appear, given the chosen language model.
Although some studies also use BLEU~\cite{papineni2002bleu} as an evaluation metric, the current literature shows that this metric is not adequate for evaluating Text Simplification~\cite{Sulem2018}.

We should note that there is a certain bias for particular types of simplifications for the SARI metric.
High SARI scores are given to simplifications that have the most number of changes~\cite{nisioi2017exploring}.

\subsection{Setup}
  
We test 8 models in this setup: 
\textit{(1)} 5 Word-Embeddings-based models (Word Embedding-based models), where the factor $\varphi$ in which the bigram perplexity affects the sentence ranking score takes the values $0$, $0.25$, $0.5$, $0.75$ and $1$, and 
\textit{(2)} 3 Transformer-based models, using the pre-trained BERT~\cite{Devlin2019bert} (bert-base-uncased), RoBERTa~\cite{Liu2019roberta} (roberta-base) and GPT2~\cite{radford2019language} (gpt2) models from \textit{Hugging Face}~\cite{Wolf2020}.

For the Word Embedding-based approach, we use Word2Vec~\cite{Mikolov2013word2vec} embeddings trained on the Wikipedia English corpus with a size of 300~\cite{word2vec}.

\subsection{Complexity prediction module results}~\label{ssec:complex-res}
    
Table~\ref{table:complex-pred} presents the evaluation of the Complexity Prediction module.
The train and test datasets are split in a 95-5 ratio, resulting 14\,250 train words and 750 test words.

    \begin{table}[!htbp]
    
    \caption{Complexity prediction system evaluation}
    \label{table:complex-pred}
    \centering
        \begin{tabular}{|l|c|c|c|}
        \hline
                \textbf{Class}    & \textbf{Precision} & \textbf{Recall} & \textbf{F1-Score} \\ \hline
        0 (simple word)  & 0.79   & 0.85     & 0.80     \\ \hline
        1 (complex word) & 0.69   & 0.64     & 0.66    \\ \hline
        \end{tabular}
    \end{table}

    The system manages to obtain good results for predicting word complexity (Table~\ref{table:complex-pred}), event though the two datasets used (i.e., the complexity ranking dataset from \cite{maddela2018word} and the news corpus from News Crawl) have never been used together.
    Words that are ranked as simple by the human volunteers have a high probability of being detected by the classification model and remain unchanged in the text.
    Thus, the Complexity Prediction module turns out to be conservative, as the evaluation shows. In other words, the model determines with a higher accuracy simple words than complex words, i.e., a precision score of 0.79 when determining simple words versus a precision of 0.69 when determining complex words.
    This is desirable, as a complexity prediction system that classifies too many words as being complex can lead to simplifications that either obfuscate the initial sentence or make it lose its meaning.
    
    To showcase the performance of our selected Multi Layer Perceptron model, we compare it with three classic machine learning models (Table~\ref{table:comparison}), i.e., Support Vector Machine, Random Forest, Extra Randomize Trees.
    For this set of experiments, we perform 10 separate runs.
    For each run, we split the dataset into 75\% training set and 25\% test set and then train and test all the considered machine learning algorithms, resulting in 11\,250 words for training and 3\,750 words for testing.
    Each dataset split uses the same label ratio as the original dataset.
    After performing the 10 runs, we compute the average and standard deviation for each metric, i.e., Accuracy, Precision, and Recall, w.r.t. the tested model.
    Among the three models, Random Forest obtains slightly better results in terms of accuracy, precision, and recall than Extra Randomize Trees.
    The Support Vector Machine obtains the overall worse results.
    Our proposed model outperforms the other models, obtaining an average accuracy of $0.79$. 
    
    \begin{table}[!htbp]
    \caption{Complexity prediction model comparison}
    \label{table:comparison}
    \centering
        \begin{tabular}{|l|c|c|c|}
        \hline
        \textbf{Model}          & \textbf{Accuracy}           & \textbf{Precision} & \textbf{Recall} \\ \hline
        Multi Layer Perceptron	& \textbf{0.79 $\pm$ 0.04}    & 0.85 $\pm$ 0.01    & 0.79 $\pm$ 0.04 \\ \hline
        Support Vector Machine	& 0.66 $\pm$ 0.01             & 0.64 $\pm$ 0.01    & 0.66 $\pm$ 0.01 \\ \hline
        Random Forest	        & 0.71 $\pm$ 0.01             & 0.71 $\pm$ 0.01    & 0.71 $\pm$ 0.01 \\ \hline
        Extra Randomize Trees	& 0.69 $\pm$ 0.01             & 0.69 $\pm$ 0.01    & 0.69 $\pm$ 0.01 \\ \hline
        \end{tabular}
    \end{table}
        
The system obtains good results, even though the two datasets used (the complexity ranking dataset from~\cite{maddela2018word} and the news corpus from News Crawl) have never been used together.
Thus, we have no guarantees that, for example, words that are ranked as simple by the human volunteers will have a high probability of apparition in the language model.
The model turns out to be a conservative one that is more confident regarding simple words than complex words.
This is desirable, as a complexity prediction system that classifies too many words as being complex can lead to simplifications that either obfuscate the initial sentence or make it lose its meaning.

Table~\ref{table:results} presents quantitative results for the task of text simplification.
Thus, we evaluate both approaches, i.e., the Word Embedding-based and Transformer-based, using SARI and Perplexity Decreases metrics.
The SARI metric quantifies both replacement correctness and syntactical simplifications for the task of text simplification.
The Perplexity Decrease shows how well the meaning is preserved while applying lexical simplifications to the original text.

 \begin{table*}[!htbp]
    \centering
    \caption{Text simplification performance results}
    \label{table:results}
        \begin{tabular}{|l|c|c|c|}
            \hline
            \textbf{Simplification model}  & \textbf{SARI} & \textbf{Perplexity Decrease} \\ \hline
            Word Embedding-based, bigram factor $\varphi = 0.00$            & 0.310 & 9.8\% \\ \hline
            Word Embedding-based, bigram factor $\varphi = 0.25$         & 0.300 & 8.4\% \\ \hline
            Word Embedding-based, bigram factor $\varphi = 0.50$          & 0.300 & 8.2\% \\ \hline
            Word Embedding-based, bigram factor $\varphi = 0.75$         & 0.300 & 7.1\% \\ \hline
            Word Embedding-based, bigram factor $\varphi = 1.00$            & 0.300 & 5.8\% \\ \hline
            Transformer-based, BERT model          & \textbf{0.350} & 8.4\% \\ \hline
            Transformer-based, RoBERTa model       & 0.349 & 8.8\% \\ \hline
            Transformer-based, GPT2 model          & 0.347 & 9.3\% \\ \hline
            NTS-w2v~\cite{nisioi2017exploring}     & 0.311 & N/A   \\ \hline
            LightLS~\cite{Glavas2015}              & 0.349 & N/A   \\ \hline
        \end{tabular}
    \end{table*}

\subsection{\textsc{SimpLex} system results}

In the experiments with Word Embedding-based approach, we use different $\varphi$ values to determine the scores change when minimizing the impact of the bigrams in the scoring function, i.e., $PP_{Score}$ (Equation~\eqref{eq:score_rw}).
We observe that after a given threshold for $\varphi$, i.e., $0.25$,  the SARI score does not change, while, by increasing the bigrams impact, the Perplexity Decrease lowers, indicating a better generalization performance.

For the Transformer-based approach, we use 3 different models: BERT, RoBERTa, and GPT3.
We observe that BERT obtains the highest SARI score, i.e., $0.350$, and Perplexity Decrease, i.e., $8.4\%$.
This indicates that the BERT transformer manages to better quantify both replacement correctness and syntactical simplifications as well as preserve the meaning, as opposed to the other 2 transformer models.

The results in Table~\ref{table:results} show that the Transformer-based approach outperforms the Word Embedding-based approach in terms of the SARI score, but in terms of Perplexity Decrease, the Word Embedding-based approach achieves the biggest decrease.
This is because the Word Embedding-based approach actively tries to find the sentence with the lowest perplexity.
We observe that the GPT2 model achieves both a good SARI score and a significant decrease in the average Perplexity of the sentences.
The Word Embedding-based models that have a high bigram factor start to suffer because the language model is not nearly big enough to capture all the bigram combinations.
In this comparison, we do not observe any important difference between the Transformer-based models. 

We compare our work with NTS-w2v~\cite{nisioi2017exploring} and LightLS~\cite{Glavas2015} (Table~\ref{table:results}).
In terms of SARI score, we observe that all the Word Embedding-based models obtain similar scores as NTS-w2v and slightly lower scores than the LightLS.
Thus, the SARI obtained with NTS-w2v is very similar to the score obtained with our Word Embedding-based approach, i.e., between 0.300 to 0.310, when varying the bigram factor $\varphi$ in range $[0.00, 1.00]$.
The Word Embedding-based approaches are outperformed by LightLS, which obtains a SARI score of 0.349 in comparison to the 0.310 obtained by the Word Embedding-based approach with a bigram factor $\varphi = 0.00$.

The Transformer-based models outperform NTS-w2v and obtain similar scores as LightLS.
With SARI scores of 0.350, 0.349, and 0.347 for the Transformer-based approaches that employ BERT, RoBERTa, and GP2, respectively, we observe that BERT obtains the best SARI score.
The Transformer-based approach that employs RoBERTa obtains the same SARI score as LightLS, i.e., 0.349, while the Transformer-based approach that employs GPT2 slightly lags behind, with a difference between the SARI scores with the best performing approach being only 0.002.

\subsection{Examples and discussion}

In this subsection, a few examples of simplification results of sentences from the employed dataset are presented.
We discuss the patterns that emerge and try to assess the simplification quality of our architecture by comparing the results obtained by \textsc{SimpLex} with the results obtained by LightLS~\cite{Glavas2015} and NTS-w2v~\cite{nisioi2017exploring}.
We present three examples in Table~\ref{tab:ex1}; examples that are discussed in detail in the following paragraphs.

\begin{table*}[!htbp]
    \tiny
    \centering
    \caption{Output comparison of the simplification system}
    \label{tab:ex1}
    \resizebox{\textwidth}{!}{%
        \begin{tabular}{ll}
            \textbf{Example~1:} & \\
            \hline
            \multicolumn{1}{|l|}{Oregano is \textbf{an} \textbf{indispensable} \textbf{ingredient} in Greek \textbf{cuisine} .} & \multicolumn{1}{l|}{Original}                                    \\ \hline
            \multicolumn{1}{|l|}{Oregano is \textbf{a} \textbf{necessary} \textbf{element} in Greek \textbf{cooking} .}         & \multicolumn{1}{l|}{Word Embedding-based, bigram factor $\varphi= 0.00$}  \\ \hline
            \multicolumn{1}{|l|}{Oregano is \textbf{a} \textbf{vital} ingredient in Greek \textbf{cooking} .}                   & \multicolumn{1}{l|}{Word Embedding-based, bigram factor $\varphi= 0.25$}  \\ \hline
            \multicolumn{1}{|l|}{Oregano is \textbf{a} \textbf{vital} ingredient in Greek \textbf{cooking} .}                   & \multicolumn{1}{l|}{Word Embedding-based, bigram factor $\varphi= 0.50$}  \\ \hline
            \multicolumn{1}{|l|}{Oregano is \textbf{a} \textbf{vital} ingredient in Greek \textbf{cooking} .}                   & \multicolumn{1}{l|}{Word Embedding-based, bigram factor $\varphi= 0.75$}  \\ \hline
            \multicolumn{1}{|l|}{Oregano is \textbf{a} \textbf{vital} ingredient in Greek cuisine .}                            & \multicolumn{1}{l|}{Word Embedding-based, bigram factor $\varphi= 1.00$}  \\ \hline
            \multicolumn{1}{|l|}{Oregano is \textbf{a} \textbf{vital} \textbf{element} in Greek \textbf{cooking} .}             & \multicolumn{1}{l|}{Transformer-based, BERT model}               \\ \hline
            \multicolumn{1}{|l|}{Oregano is \textbf{a} \textbf{critical} \textbf{base} in Greek \textbf{cooking} .}             & \multicolumn{1}{l|}{Transformer-based, RoBERTa model}            \\ \hline
            \multicolumn{1}{|l|}{Oregano is \textbf{a} \textbf{vital} \textbf{element} in Greek \textbf{preparation} .}         & \multicolumn{1}{l|}{Transformer-based, GPT2 model}               \\ \hline
            \multicolumn{1}{|l|}{Oregano is \textbf{an} \textbf{essential} ingredient in Greek cuisine .}                       &  \multicolumn{1}{l|}{LightLS~\cite{Glavas2015}}                  \\ \hline
            \multicolumn{1}{|l|}{Oregano is \textbf{a} \textbf{vital} \textbf{element} in Greek cuisine .}                      & \multicolumn{1}{l|}{NTS-w2v~\cite{nisioi2017exploring}}          \\ \hline
            \hline \\
            \textbf{Example~2:}\label{texample3} & \\ \hline
            \multicolumn{1}{|l|}{It is \textbf{situated} \textbf{at} the coast of the Baltic Sea , where it \textbf{encloses} the city of Stralsund .} & \multicolumn{1}{l|}{Original}                                   \\ \hline
            \multicolumn{1}{|l|}{It is \textbf{find out} \textbf{at} the coast of the Baltic Sea , where it \textbf{tubes} the city of Stralsund .}    & \multicolumn{1}{l|}{Word Embedding-based, bigram factor $\varphi= 0.00$}    \\ \hline
            \multicolumn{1}{|l|}{It is \textbf{find out} \textbf{at} the coast of the Baltic Sea , where it \textbf{wraps} the city of Stralsund .}    & \multicolumn{1}{l|}{Word Embedding-based, bigram factor $\varphi= 0.25$} \\ \hline
            \multicolumn{1}{|l|}{It is \textbf{find out} \textbf{at} the coast of the Baltic Sea , where it \textbf{wraps} the city of Stralsund .}    & \multicolumn{1}{l|}{Word Embedding-based, bigram factor $\varphi= 0.50$}  \\ \hline
            \multicolumn{1}{|l|}{It is \textbf{based} \textbf{at} the coast of the Baltic Sea , where it \textbf{wraps} the city of Stralsund .}       & \multicolumn{1}{l|}{Word Embedding-based, bigram factor $\varphi= 0.75$} \\ \hline
            \multicolumn{1}{|l|}{It is \textbf{find out} \textbf{at} the coast of the Baltic Sea , where it \textbf{wraps} the city of Stralsund .}    & \multicolumn{1}{l|}{Word Embedding-based, bigram factor $\varphi= 1.00$}    \\ \hline
            \multicolumn{1}{|l|}{It is \textbf{located} \textbf{at} the coast of the Baltic Sea , where it \textbf{covers} the city of Stralsund .}    & \multicolumn{1}{l|}{Transformer-based, BERT model}              \\ \hline
            \multicolumn{1}{|l|}{It is \textbf{determined} \textbf{at} the coast of the Baltic Sea , where it \textbf{bathes} the city of Stralsund .} & \multicolumn{1}{l|}{Transformer-based, RoBERTa model}           \\ \hline
            \multicolumn{1}{|l|}{It is \textbf{determined} \textbf{at} the coast of the Baltic Sea , where it \textbf{tubes} the city of Stralsund .}  & \multicolumn{1}{l|}{Transformer-based, GPT2 model}              \\ \hline
            \multicolumn{1}{|l|}{It is \textbf{near} \textbf{at} the coast of the Baltic Sea , where it \textbf{surrounds} the city of Stralsund .}    & \multicolumn{1}{l|}{LightLS~\cite{Glavas2015}}                 \\ \hline
            \multicolumn{1}{|l|}{It is \textbf{located} \textbf{on} the coast of the Baltic Sea , where it \textbf{surrounds} the town of Stralsund .} & \multicolumn{1}{l|}{NTS-w2v~\cite{nisioi2017exploring}}         \\ \hline
            \hline \\
            \textbf{Example~3:} & \\ \hline
            \multicolumn{1}{|l|}{Since 2000 , the \textbf{recipient} of the Kate Greenaway Medal has also been \textbf{awarded} the £5,000 Colin Mears Award .} & \multicolumn{1}{l|}{Original}                                   \\ \hline
            \multicolumn{1}{|l|}{Since 2000 , the \textbf{host} of the Kate Greenaway Medal has also been awarded the £5,000 Colin Mears Award .}               & \multicolumn{1}{l|}{Word Embedding-based, bigram factor $\varphi= 0.00$} \\ \hline
            \multicolumn{1}{|l|}{Since 2000 , the \textbf{host} of the Kate Greenaway Medal has also been awarded the £5,000 Colin Mears Award .}               & \multicolumn{1}{l|}{Word Embedding-based, bigram factor $\varphi= 0.25$} \\ \hline
            \multicolumn{1}{|l|}{Since 2000 , the \textbf{host} of the Kate Greenaway Medal has also been awarded the £5,000 Colin Mears Award .}               & \multicolumn{1}{l|}{Word Embedding-based, bigram factor $\varphi= 0.50$} \\ \hline
            \multicolumn{1}{|l|}{Since 2000 , the \textbf{host} of the Kate Greenaway Medal has also been awarded the £5,000 Colin Mears Award .}               & \multicolumn{1}{l|}{Word Embedding-based, bigram factor $\varphi= 0.75$} \\ \hline
            \multicolumn{1}{|l|}{Since 2000 , the \textbf{host} of the Kate Greenaway Medal has also been awarded the £5,000 Colin Mears Award .}               & \multicolumn{1}{l|}{Word Embedding-based, bigram factor $\varphi= 1.00$} \\ \hline
            \multicolumn{1}{|l|}{Since 2000 , the \textbf{heir} of the Kate Greenaway Medal has also been awarded the £5,000 Colin Mears Award .}               & \multicolumn{1}{l|}{Transformer-based, BERT model}              \\ \hline
            \multicolumn{1}{|l|}{Since 2000 , the \textbf{dependent} of the Kate Greenaway Medal has also been awarded the £5,000 Colin Mears Award .}          & \multicolumn{1}{l|}{Transformer-based, RoBERTa model}           \\ \hline
            \multicolumn{1}{|l|}{Since 2000 , the \textbf{receiver} of the Kate Greenaway Medal has also been awarded the £5,000 Colin Mears Award .}           & \multicolumn{1}{l|}{Transformer-based, GPT2 model}              \\ \hline
            \multicolumn{1}{|l|}{Since 2000 , the recipient of the Kate Greenaway Medal has also been \textbf{received} the £5,000 Colin Mears Award .}         & \multicolumn{1}{l|}{LightLS~\cite{Glavas2015}}                  \\ \hline
            \multicolumn{1}{|l|}{Since 2000 , the \textbf{host} of the Kate Greenaway Medal has also been \textbf{made} with the £5,000 Colin Mears Award .}    & \multicolumn{1}{l|}{NTS-w2v~\cite{nisioi2017exploring}}         \\ \hline
        \end{tabular}%
    }
    \end{table*}

For the first example (Example 1 from Table~\ref{tab:ex1}), the simplification system has selected 3 words as being candidates for simplification, i.e., \textit{indispensable}, \textit{ingredient}, and \textit{cooking}.
Both the Transformer and the Word Embedding-based models perform well, finding suitable replacements for the given words.
As the bigram factor increases in the Word Embedding-based models, fewer words are replaced.
Interestingly, there are some visible differences between the chosen words among the transformer models.
Among those, a visual examination tends to suggest that the BERT model achieves the most reasonable result, while the other two models tend to judge the context slightly poorly.
We observe that LightLS~\cite{Glavas2015} changes \textit{indispensable} with \textit{essential}.
For \textit{indispensable ingredient}, TS-w2v~\cite{nisioi2017exploring} does the same replacement as our Transformer-based model that employs BERT.
Both LightLS~\cite{Glavas2015} and TS-w2v~\cite{nisioi2017exploring} do not manage to replace the complex word \textit{cuisine}.

For the second example (Example 2 from Table~\ref{tab:ex1}), we observe the improvements made by the Transformer-based models in comparison with the Word Embedding-based models.
In the Word Embedding-based models, the words \textit{situated} and \textit{encloses} are replaced by sub-optimal candidates, as the model has no knowledge of the context in which the words are used.
The context refers to a geographical location, so the most appropriate words to be used are those that refer to spatial placement.
The transformer models do not suffer from this drawback and manage to find more suitable and even natural replacements for the selected words.
There are, again, differences between the models.
RoBERTa and GPT2 choose strange candidates for the word \textit{encloses}, while BERT seems to choose the most natural replacement among the tested models.
Both LightLS~\cite{Glavas2015} and NTS-w2v~\cite{nisioi2017exploring} replace \textit{encloses} with \textit{surrounds}, while for \textit{situated}, LightLS~\cite{Glavas2015} uses \textit{near} with the wrong preposition \textit{at} and NTS-w2v~\cite{nisioi2017exploring} uses \textit{located on}.

In the last example (Example 3 from Table~\ref{tab:ex1}), we present a single word replacement, i.e., \textit{recipient}, in the context of a larger simple sentence.
The larger context refers to a medal nomination, so we anticipate the perfect candidate to be along the line of \textit{receiver} or \textit{winner}.
As it can be seen, the Word Embedding-based models fail to capture the context, as expected, and produce a sub-par simplification, choosing the word \textit{host}.
The choice is not satisfactory, as it changes the meaning of the sentence.
Now, the subject of the sentence refers no more to the recipient of the medal but the host that awards the medal.
This is a perfect example in which the lack of context can significantly hurt the performance of the simplification architecture.
The Transformer-based models benefit from context awareness and have a better chance of finding the right candidate.
Thus, among the three pre-trained models tested, BERT and RoBERTa find sub-par, strange candidates, but the GPT2 model finds a good match in the word \textit{receiver}.
For this example, LightLS~\cite{Glavas2015} does not replace \textit{recipient}, while NTS-w2v~\cite{nisioi2017exploring} replaces it with \textit{host}.
Although \textsc{SimpLex} classifies \textit{awarded} as a simple word, both LightLS~\cite{Glavas2015} and NTS-w2v~\cite{nisioi2017exploring} replace it with \textit{received} and \textit{made}, respectively.

\section{Conclusions}~\label{sec:conclusions}

In this paper, we present \textsc{SimpLex}, a novel lexical simplification architecture that employs both word and transformers embeddings --- achieving objective $O_1$.
\textsc{SimpLex} uses either a Word Embedding-based or a Transformer-based approach to generate simplified sentences --- answering the research question $Q_1$.
The Word Embedding-based approach uses Word2Vec and perplexity, while the Transformer-based approach uses three transformers, i.e., BERT, RoBERTa, and GPT2, and cosine similarity.
We perform ample experiments to show the feasibility of our architecture.
For evaluation, we use two metrics, i.e., SARI and Perplexity Decrease.
We compare our solution with two state-of-the-art models, i.e.,  LightLS~\cite{Glavas2015} and NTS-w2v~\cite{nisioi2017exploring} --- achieving objective $O_1$.
We conclude that the Transformer-based approach is more suited for the task of text simplification as transformer word and sentence embeddings better preserve the context improving the task of synonym detection and should be used together.

Furthermore, \textsc{SimpLex} provides a simple-to-use and friendly user interface --- answering the research question $Q_2$.
It can be run either from the command line or as a docker.
We also provide the code for further development for interested users and researchers in the field of text simplification.

The current research identifies a series of shortcomings in the task of designing and developing text simplification systems~\cite{Stajner2021}, such as:
\textit{(1)} publicly available datasets for low resource languages,
\textit{(2)} evaluation metrics that are focused on the final user,
\textit{(3)} quality of preserving grammaticality and meaning by the automatic text simplification systems, and
\textit{(4)} simple to use systems by non-specialized users.

The research community is trying to address some of these shortcomings by proposing new research tasks and directions~\cite{Ermakova2021}:
\textit{(1)} word or sentence ranking,
\textit{(2)} background knowledge searching,
\textit{(3)} text simplification for scientific datasets
With \textsc{SimpLex}, our proposed novel text simplification system, we try to address some of these shortcomings.
To summarize our contributions:
\begin{itemize}
    \item [\textit{(1)}] We propose a new algorithm for text simplification that determines the complexity of words and utilizes Word Embedding-based or Transformer-based approach to generate simplified sentences;
    \item [\textit{(2)}] We present a novel text simplification system called \textsc{SimpLex} that offers an intuitive user interface and a modular design that can help future development as well as provide a baseline for further research;
    \item [\textit{(3)}] We perform an in-depth analysis of our solution and compare our results with two state-of-the-art models, i.e.,  LightLS~\cite{Glavas2015} and NTS-w2v~\cite{nisioi2017exploring}.    
\end{itemize}

In future work, we aim to add a syntactic simplification module.
We also plan to use graph embedding and word representation formalism, e.g., Abstract Meaning Representation (AMR)~\cite{Banarescu2013} or MRS (Minimal Recursion Semantics)~\cite{Copestake2005}, for text simplification.

\section*{Compliance with ethical standards}

\textbf{Conflict of interest} The authors declare that they have no conflict of interest.

\bibliographystyle{plainnat}
\bibliography{main}

\end{document}